# Counterfactuals and Policy Analysis in Structural Models

**Alexander Balke**
Cognitive Systems Laboratory
Computer Science Department
University of California
Los Angeles, CA 90024

**Judea Pearl**
Cognitive Systems Laboratory
Computer Science Department
University of California
Los Angeles, CA 90024

## Abstract

Evaluation of counterfactual queries (e.g., "If $A$ were true, would $C$ have been true?") is important to fault diagnosis, planning, determination of liability, and policy analysis. We present a method for evaluating counterfactuals when the underlying causal model is represented by structural models – a nonlinear generalization of the simultaneous equations models commonly used in econometrics and social sciences. This new method provides a coherent means for evaluating policies involving the control of variables which, prior to enacting the policy were influenced by other variables in the system.

## 1 INTRODUCTION

A counterfactual sentence has the form

*If $A$ were true, then $C$ would have been true*

where $A$, the counterfactual antecedent, specifies an event that is contrary to one's real-world observations, and $C$, the counterfactual consequent, specifies a result that is expected to hold in the alternative world where the antecedent is true. A typical example is "If Oswald were not to have shot Kennedy, then Kennedy would still be alive," which presumes the factual knowledge that Oswald did shoot Kennedy, contrary to the antecedent of the sentence.

Counterfactual reasoning is at the heart of every planning activity, especially real-time planning. When a planner discovers that the current state of affairs deviates from the one expected, a "plan repair" activity need be invoked to determine what went wrong and how it could be rectified. This activity amounts to an exercise of counterfactual thinking, as it calls for rolling back the natural course of events and determining, based on the factual observations at hand, whether the culprit lies in previous decisions or in some unexpected, external eventualities. Moreover, in reasoning forward to determine if things would have been different a new model of the world must be consulted, one that embodies hypothetical changes in decisions or eventualities, hence, a breakdown of the old model or theory.

The logic-based planning tools used in AI, such as STRIPS and its variants or those based on the situation calculus, do not readily lend themselves to counterfactual analysis; as they are not geared for coherent integration of abduction with prediction, and they do not readily handle theory changes. Remarkably, the formal system developed in economics and social sciences under the rubric "structural equations models" does offer such capabilities but, as will be discussed below, these capabilities are not well recognized by current practitioners of structural models.[1] The purpose of this paper is both to illustrate to AI researchers the basic formal features needed for counterfactual and policy analysis, and to call the attention of economists and social scientists to capabilities that are dormant within structural equations models.

Counterfactual thinking dominates reasoning in political science and economics. We say, for example, "If Germany were not punished so severely at the end of World War I, Hitler would not have come to power," or "If Reagan did not lower taxes, our deficit would be lower today." Such thought experiments emphasize an understanding of generic laws in the domain and are aimed toward shaping future policy making, for example, "defeated countries should not be humiliated," or "lowering taxes (contrary to Reaganomics) tends to increase national debt."

Strangely, there is very little formal work on counterfactual reasoning or policy analysis in the behavioral science literature. An examination of a number of econometric journals and textbooks, for example, reveals an imbalance: while an enormous mathemat-ical machinery is brought to bear on problems of estimation and prediction, policy analysis (which is the ultimate goal of economic theories) receives almost no formal treatment. Currently, the most popular methods driving economic policy making are based on so-called *reduced-form* analysis: to find the impact of a policy involving decision variables $X$ on outcome variables $Y$, one examines past data and estimates the conditional expectation $E(Y|X=x)$, where $x$ is the particular instantiation of $X$ under the policy studied.

---

[1] These were clearly recognized though by the founding fathers of structural models, as can be seen in the publications of the Cowels Commission [Haavelmo, 1943] [Marschak, 1950] [Simon, 1953] but, with the exception of [Strotz and Wold, 1971], [Simon and Rescher, 1966] and [Fisher, 1970], have all but disappeared from the econometrics literature.

The assumption underlying this method is that the data were generated under circumstances in which the decision variables $X$ act as exogenous variables, that is, variables whose values are determined outside the system under analysis. However, while new decisions should indeed be considered exogenous for the purpose of evaluation, past decisions are rarely enacted in an exogenous manner.[2] Almost every realistic policy (e.g., taxation) imposes control over some endogenous variables, that is, variables whose values are determined by other variables in the analysis. Let us take taxation policies as an example. Economic data are generated in a world in which the government is reacting to various indicators and various pressures; hence, taxation is endogenous in the data-analysis phase of the study. Taxation becomes exogenous when we wish to predict the impact of a specific decision to raise or lower taxes. The reduced-form method is valid only when past decisions are nonresponsive to other variables in the system, and this, unfortunately, eliminates most of the interesting control variables (e.g., tax rates, interest rates, quotas) from the analysis.[3]

---

[2]This distinction is often blurred in the literature. [Druzdzel and Simon, 1993], for example, state: "A variable is considered exogenous to a system if its value is determined outside the system, either because we can control its value externally (e.g., the amount of taxes in a macroeconomic model) or because we believe that this variable is controlled externally (like the weather in a system describing crop yields, market prices, etc.)" Still, our ability to externally control the value of a variable $X$ does not render $X$ exogenous for the purpose of legitimizing the reduced form analysis: for $E[Y|X=x]$ to represent the impact of $X=x$ on $Y$, $X$ must also be independent of all implicit factors (disturbance terms) affecting $Y$.

While every economist knows that this disturbance-independence is a necessary condition for consistent estimation of structural parameters, most economists assume that disturbance-independence is a guaranteed property of controllable policy variables. A popular textbook [Intriligator, 1978], for example, mentions these two properties as if they were synonymous: "The exogenous variables are variables the values for which are determined outside the model but which influence the model. From a formal standpoint the exogenous variables are assumed to be statistically independent of all stochastic disturbance terms of the model, while the endogenous variables are not statistically independent of those terms. ... In general the exogenous variables are either historically given, policy variables, or determined by some separate mechanism."

[3]This problem is unrelated to the celebrated Lucas's critique [Lucas, 1976] which concerns parameter changes

This difficulty is not unique to economic or social policy making; it appears whenever one wishes to evaluate the merit of a plan on the basis of the past performance of other agents. Even when the signals triggering the past actions of those agents are known with certainty, a systematic method must be devised for selectively ignoring the influence of those signals from the evaluation process. In fact, the very essence of *evaluation* is having the freedom to imagine and compare trajectories in various counterfactual worlds, where each world or trajectory is created by a hypothetical implementation of a policy that is free of the very pressures that compelled the implementation of such policies in the past.

A connection between counterfactuals and policy making was formulated in [Balke and Pearl, 1994b] using a simple device from action theory. In that formulation, the counterfactual antecedent is interpreted as a hypothetical minimal intervention that forces the counterfactual antecedent to hold true. If a system is modeled with structural equations (respectively, causal graphs), an intervention is simulated by severing all equations (causal edges) that correspond to (lead into) the antecedent variables and setting their values to those specified in the antecedent [Strotz and Wold, 1971]. A calculus for working with interventions in causal theories is given in [Pearl, 1994].

[Balke and Pearl, 1994b] provides background and motivation for the evaluation of counterfactual conditionals and briefly illustrates how the intervention scheme would handle counterfactuals in models represented by linear structural equations. This paper amends and expands the treatment of counterfactuals and policy making in any structural model for which the form of the equations is give. It also presents an example of their use in the area of econometrics, where apparently no adequate formalism for dealing with policy analysis has been proposed. In contrast to reduced-form analysis, our method allows evaluation of the consequences of intervening on economic attributes that are endogenous in normal operation only to become exogenous for the purpose of evaluation. For example, after developing the general techniques in Section 3, we will illustrate their use in Section 4 by evaluating the effect on the demand for some commodity when a government imposes price controls on that commodity for the first time.

## 2   REVIEW OF COUNTERFACTUAL ANALYSIS

In this section, the procedure for evaluating counterfactual conditionals in the context of structural equa-

---

due to economic agents becoming aware of interventions. The failure of reduced-form analysis extends to physical systems as well, where there are no rational agents to speak of, and where system parameters remain unaltered (except those under direct control).

tion models will be reviewed. We will then demonstrate this procedure on an example where the relationships among observed variables are deterministic, followed by an example that demonstrates how exceptions and disturbances incorporated into the model affect the analysis procedure.

Let $V = \{V_1, V_2, \ldots, V_n\}$ represents the set of variables for which data may be observed in a system. $U = \{U_1, U_2, \ldots, U_m\}$ will represent the disturbances, exceptions, and/or abnormalities influencing the observable variables $V$. For example, $U$ could summarize the influence of many exogenous factors, such as the "price of tea in China" or "the local weather." In general, each observable variable $V_i$ is a deterministic function of the form:

$$V_i = f_i(V_1, V_2, \ldots, V_n, U_1, U_2, \ldots, U_m) \quad (1)$$

The structure of the model defined by these equations may be depicted by a causal graph, where each variable on the left hand side of a structural equation is the child of those variables on the right hand side of the equation. A probability distribution over the disturbances, $P(u_1, u_2, \ldots, u_m)$, embodies the nondeterminism in the model. In general, this distribution is unconstrained; however, some classes of models, e.g., regression models, will assume that the disturbance variables $U_k$ are mutually independent.

A counterfactual conditional will be written

$$a \rightarrow c \mid o \quad (2)$$

and read as "Given that we have observed $o$, if $a$ were true, then $c$ would have been true." The observations $o$ consists of a set of value assignments to variables in $V$, e.g., $V_j = v_j, V_k = v_k$. The counterfactual antecedent $a$, consists of a conjunction of value assignments to variables in $V$ that are forced to hold true by external intervention. Typically, to justify being called "counterfactual", $a$ conflicts with $o$. Finally, the counterfactual consequent, $c$, stands for the proposition of interest, usually the values attained by some variables in the system.

The truth of a counterfactual conditional $a \rightarrow c \mid o$ may then be evaluated by the following procedure:

- Use the observations $o$ to update the joint belief[4] for all root nodes in the causal network. This joint belief summarizes the state of the system, because each non-root variable is a deterministic function of its causal influences.

- Replace the structural equation for each variable $V_k$ referred to in the counterfactual antecedent $a$ with the equation $V_k = a_{v_k}$ where $a_{v_k}$ is the value of $V_k$ specified in $a$. This implements the local intervention that forces the counterfactual antecedent to hold true.

---
[4]Here we use the generic term "belief" to refer to either truth assignments or probabilities.

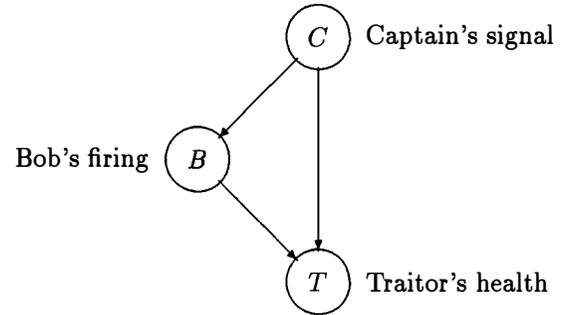

Figure 1: *Causal structure reflecting the influence that the Captain's signal has on Bob's firing and the Traitor's health, and the direct influence that Bob's firing has on the Traitor's health.*

- Compute the solutions or belief of the consequent proposition $c$ according to the modified set of structural equations.

This procedure will work whenever we have the functional form of the $f_i$'s, in which case the model is called *parametric*; otherwise, the model is called *nonparametric*. In particular, this paper concentrates on linear and boolean functions (e.g., Noisy-OR gates). In the case that the model is nonparametric, only bounds may be calculated for the belief of a counterfactual consequent [Balke and Pearl, 1994a].

To illustrate the intervention-based interpretation of counterfactuals, consider a firing squad with several riflemen (one called Bob) and a Captain who gives a signal to either shoot or release a prisoner charged with treason. The behavior of these agents is as follows:

- The Captain waits for the court decision.
- Bob typically fires his rifle if and only if the Captain gives the signal to shoot.
- The Traitor typically dies if and only if the Captain gives the signal to shoot or Bob fires his rifle.

Note that if the Captain gives the signal to shoot and Bob does not fire, the traitor will typically die as a result of the other riflemen shooting, but these intermediate causes will not be made explicit in this story in order to keep the model simple.

The generic causal structure that reflects this description is represented in Figure 1. The three variables $C$, $B$, and $T$ have the following domains:

$$c \in \left\{\begin{array}{l} 0 \equiv \text{Captain gives the signal to release the traitor.} \\ 1 \equiv \text{Captain gives the signal to shoot the traitor.} \end{array}\right\}$$

$$b \in \left\{\begin{array}{l} 0 \equiv \text{Bob does not fire his rifle.} \\ 1 \equiv \text{Bob fires his rifle.} \end{array}\right\}$$

$$t \in \left\{\begin{array}{l} 0 \equiv \text{Traitor lives.} \\ 1 \equiv \text{Traitor dies.} \end{array}\right\}$$

The following subsections will demonstrate the evaluation of counterfactual conditionals under two varia-tions of this model. The first assumes that the behaviors of the characters in the story are deterministic, while the second admits the occurrence of exceptions.

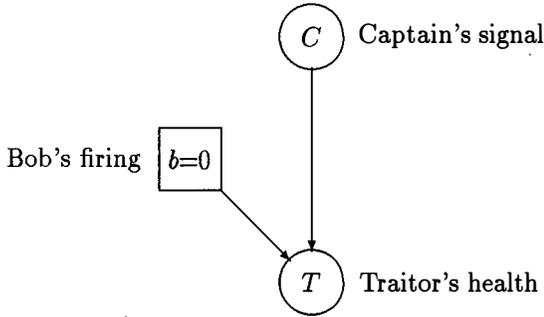

Figure 2: *Causal structure reflecting an external intervention that forces the state of Bob's firing despite its normal causal influences, e.g., the Captain's signal.*

## 2.1 DETERMINISTIC ANALYSIS

A deterministic model is a special case of the general structural equation model where the disturbance variables set the values of the root nodes in the causal graph, i.e.,

$$V_i = f_i(U_k) \qquad (3)$$

and the remaining observable variables (those that are not root nodes) are deterministic functions of the set of observable variables $V$, i.e.,

$$V_i = f_i(V_1, V_2, \ldots, V_n) \qquad (4)$$

In a deterministic model, the firing-squad story may be concisely expressed by the following structural equations

$$B = C \qquad (5)$$
$$T = B \vee C \qquad (6)$$

Suppose that we observe Bob fire his rifle ($b = 1$) and the traitor expires ($t = 1$). If Bob were not to have fired ($b = 0$), would the traitor have lived ($t = 0$), i.e., does $b=0 \to t=0 \mid b=1, t=1$ hold true? Following the procedure previously outlined, the belief in the root nodes of the causal structure are first evaluated; in this case, did the captain give the order to fire? Applying Eq. (5) allows us to abductively infer that the Captain must have given the order to fire ($c = 1$).

The structural equations (and hence the causal structure) are then modified to reflect an external intervention forcing Bob to have not fired ($b = 0$):

$$B = 0 \qquad (7)$$
$$T = B \vee C \qquad (8)$$

Figure 2 depicts the causal structure reflecting this modified set of structural equations.

Finally, substituting our previously computed beliefs for the root nodes in the causal structure, i.e., that the Captain gave the order to fire ($c = 1$), evaluate our belief in the traitor's state of health. In our example query, substitute $c = 1$ and $b = 0$ from the intervention into Eq. (8) to conclude that the traitor would still have died ($t = 1$). Therefore, the analysis leads to the statement, "given that Bob fired his rifle and the traitor died, if Bob had not fired his rifle, the traitor would still have died."

This method for analyzing counterfactual conditionals was developed with the goal of preventing reasoning from the counterfactual antecedent variables to their ancestors in the causal structure, e.g., to conclude that the Captain would not have given the signal to shoot, if Bob did not fire. Such abductive reasoning is legitimate in an unchanged, typical world but does not reflect the subjucntive mood of the counterfactual which invites unexpected eventualities (e.g., Bob failing to or deciding not to fire), similar to eventualities that are considered in decision making.

This solution is essentially the same as would be computed by [Simon and Rescher, 1966], who suppress abductive inference by invoking only forward inferences. Our method, which suppresses abduction by removing equations from the model, has two advantages. In the probabilistic analysis, our method permits the counterfactual computation using ordinary evidence propagation in a dual network [Balke and Pearl, 1994b]. Moreover, our proposal is also applicable to nonrecursive theories as will be shown in Section 3.

## 2.2 ASSUMPTION-BASED ANALYSIS

In the previous subsection we assumed that there were no exceptions to the normal behaviors of each of the characters in the story. A more realistic model of the story would be to incorporate assumptions and exceptions that effect how each observable variable is effected by its observable causal influences. For example, in the firing-squad story, there may be exceptions to Bob's firing his rifle in accordance with the Captain's signal: his rifle may become jammed preventing him from firing, or he may have had an itchy trigger finger. In addition, the traitor may have a cardiac arrest and die without anyone firing, or all the riflemen may miss the target. In order to accommodate these eventualities without explicating every possible scenario, we will write the structural equations with exception terms:

$$B = (C \vee ab_{b1}) \wedge \neg ab_{b2} \qquad (9)$$
$$T = (B \vee C) \wedge \neg ab_{t1} \vee ab_{t2} \qquad (10)$$

$ab_{b1}$ summarizes events that can cause Bob to fire even though the Captain did not give the order to fire, while $ab_{b2}$ summarizes those events that can prevent Bob from firing his rifle. Likewise, $ab_{t2}$ summarizes those events that can cause the Traitor to die even though Bob did not fire and the Captain did not give the order to fire, while $ab_{t1}$ summarizes those events that can prevent the Traitor from expiring even though the riflemen fired. These abnormality variables correspond to the set of disturbance variables $U$ described in our definition of structural equations models.

If we apply the previous query to this assumption-based model, the same conclusion will be obtained, because the most believable world consistent with the observations contains no exceptions, which reduces Eqs. (9) and (10) to Eqs. (5) and (6). Therefore, we will work on a more complex query where abnormalities make a difference in the conclusion. Suppose that we observe the Captain give the signal to release the traitor ($c = 0$) and the Traitor expires ($t = 1$). Given this data there is a possibility that Bob's firing was an accident. Now we ask: If Bob were not to have fired ($b = 0$), would the Traitor have lived ($t = 0$)[5], i.e., does $b = 0 \rightarrow t = 0 \mid c = 0, t = 1$ hold true? As before, we compute updated beliefs for each root variable in the model given the observations. Our belief in $C$ is already given by the observation, so we only need to compute our belief in the abnormality variables, e.g., $ab_{b1}$.

Qualitatively, those states of the world that minimize the number of abnormalities (exceptions) are to be assigned the highest belief. The fact that the Captain gave the release signal and the Traitor expired, tells us that there is at least one abnormal condition. Indeed, there are exactly two assignments to the root variables that satisfy the observations and only contain one abnormal condition:

$$(C = 0, ab_{b1} = 1, ab_{b2} = 0, ab_{t1} = 0, ab_{t2} = 0) \quad (11)$$
$$(C = 0, ab_{b1} = 0, ab_{b2} = 0, ab_{t1} = 0, ab_{t2} = 1) \quad (12)$$

The effect of the external intervention that forces Bob not to fire his rifle is to be computed under these two states of the system. First the structural equations are modified to reflect the external intervention:

$$B = 0 \quad (13)$$
$$T = (B \vee C) \wedge \neg ab_{t1} \vee ab_{t2} \quad (14)$$

Substituting the values from Eq. (11) into these equations leads to the belief that the Traitor would be alive. Intuitively, this particular state corresponds to the case where Bob had an itchy trigger finger and hence killed the Traitor; if Bob were prevented from firing, the mechanism responsible for the Traitor's death is disabled and the Traitor would have lived.

However, substituting the values from Eq. (12) into the revised structural equations leads to the alternative conclusion that the Traitor would still have died. In this state, the Traitor died from fright, and would have expired even if Bob were prevented from firing.

If the exceptions represented by $ab_{b1}$ are more likely than the exceptions represented by $ab_{t2}$, then we would choose to believe that the Traitor would have lived. Otherwise, we would conclude that the Traitor would still have expired.

## 3 LINEAR-NORMAL MODELS

The remainder of the paper will concentrate on models where the functions of Eq. (1) are linear and the disturbances are normally distributed. Some notation will be helpful for expressing background knowledge and counterfactual queries in this class of models. Upper case letters (e.g., $Q$) represent variables and the corresponding lower case letters (e.g., $q$) represent the value of those variables. When referring to a set of variables or values, we will use vector notation (e.g., $\vec{X}$ and $\vec{x}$); however, the arrow will be dropped whenever the variable is used as a subscript and its context is known. The distribution of variables in a linear structural equation model with Gaussian disturbances is fully specified by a mean vector ($\vec{\mu}_x$) and a covariance matrix ($\Sigma_{x,x}$).

Counterfactual distributions will be notated by $\mu_{c^*|\hat{a}^*,o}$ and $\Sigma_{c^*,c^*|\hat{a}^*,o}$ which may be read as the "mean and covariance of $c$ given the observations $o$, if $a$ were true (counterfactually)."

Assume that knowledge is specified by the linear structural equation model (often used in econometrics and the social sciences, and originally established by Sewall Wright in his development of path analysis [Wright, 1921])

$$\vec{x} = B\vec{x} + \vec{\epsilon}$$

where $B$ is a matrix (not necessarily triangular) corresponding to a causal model (possibly cyclic), and we are given the mean $\vec{\mu}_\epsilon$ and covariance $\Sigma_{\epsilon,\epsilon}$ of the disturbances $\vec{\epsilon}$ (assumed to be normal). The variables on the right-hand side of a structural equation are interpreted as the causal influences of the variable on the left-hand side of the equation. The mean and covariance of the observable variables $\vec{X}$ are then given by

$$\vec{\mu}_x = S\vec{\mu}_\epsilon \quad (15)$$
$$\Sigma_{x,x} = S\Sigma_{\epsilon,\epsilon}S^t \quad (16)$$

where $S = (I - B)^{-1}$.

Under such a model, there are well-known formulas [Whittaker, 1990, p. 163] for evaluating the mean and covariance of $\vec{X}$ conditioned on some observations $\vec{o}$:

$$\vec{\mu}_{x|o} = \vec{\mu}_x + \Sigma_{x,o}\Sigma_{o,o}^{-1}(\vec{o} - \vec{\mu}_o) \quad (17)$$
$$\Sigma_{x,x|o} = \Sigma_{x,x} - \Sigma_{x,o}\Sigma_{o,o}^{-1}\Sigma_{o,x} \quad (18)$$

where, for every pair of subvectors, $\vec{Z}$ and $\vec{W}$, of $\vec{X}$, $\Sigma_{z,w}$ is the submatrix of $\Sigma_{x,x}$ with entries corresponding to the components of $\vec{Z}$ and $\vec{W}$. Singularities of $\Sigma$ terms are handled by appropriate means.

Similar formulas apply for the mean and covariance of $\vec{X}$ under an action $\tilde{a}$. For mathematical convenience,

---

[5]This counterfactual conditional differs from most in that no direct observation has been made for the variable referred to in the counterfactual antecedent; hence, technically, the conditional may or may not be "counterfactual." The interpretation of a local-intervention on the antecedent variable, though, is still clear, and the analysis procedure can compute a meaningful belief for the counterfactual consequent.

let $\vec{X}$ be partitioned according to whether each variable is referred to in $\vec{a}$. The set of variables referred to in $\vec{a}$ is denoted by $\vec{Z}$, and the set of remaining variables in $\vec{X}$ is denoted by $\vec{Y}$. Under this partition, the matrix $B$ can be partitioned into four submatrices:

$$B = \begin{bmatrix} B_{yy} & B_{yz} \\ B_{zy} & B_{zz} \end{bmatrix}$$

$B$ is replaced by the action-pruned matrix $\hat{B} = [\hat{b}_{ij}]$, defined by

$$\hat{b}_{ij} = \begin{cases} 0 & \text{if } X_i \in \vec{a} \\ b_{ij} & \text{otherwise} \end{cases}$$

Equivalently,

$$\hat{B} = \begin{bmatrix} B_{yy} & B_{yz} \\ 0 & 0 \end{bmatrix}$$

According to intervention semantics [Pearl, 1994], all links from $\vec{\epsilon}_z$ to $\vec{Z}$ are severed and $\vec{Z}$ is forced to the value $\vec{a}$. Therefore, the modified structural equation model for $\vec{X}$ when influenced by external actions is given by

$$\vec{x} = (I - \hat{B})^{-1} \begin{bmatrix} \vec{\epsilon}_y \\ 0 \end{bmatrix} + (I - \hat{B})^{-1} \begin{bmatrix} 0 \\ \vec{a} \end{bmatrix}$$

Given the mean and covariance of $\vec{\epsilon}_y$, the mean and covariance of the observable variables $\vec{X}$ may be evaluated:

$$\vec{\mu}_{x|\hat{a}} = \begin{bmatrix} \vec{\mu}_{y|\hat{a}} \\ \vec{\mu}_{z|\hat{a}} \end{bmatrix}$$

$$= \begin{bmatrix} (I - B_{yy})^{-1}(\vec{\mu}_{\epsilon_y} + B_{yz}\vec{a}_z) \\ \vec{a}_z \end{bmatrix} \quad (19)$$

$$\Sigma_{x,x|\hat{a}} = \Sigma_{yz,yz|\hat{a}}$$

$$= \begin{bmatrix} \Sigma_{y,y|\hat{a}} & \Sigma_{y,z|\hat{a}} \\ \Sigma_{z,y|\hat{a}} & \Sigma_{z,z|\hat{a}} \end{bmatrix}$$

$$= \begin{bmatrix} (I - B_{yy})^{-1}\Sigma_{\epsilon_y,\epsilon_y}((I - B_{yy})^{-1})^t & 0 \\ 0 & 0 \end{bmatrix} (20)$$

To evaluate the counterfactual distribution $\mu_{x^*|\hat{a}^* o}$ and $\Sigma_{x^*,x^*|\hat{a}}$, we first update the prior distribution of the disturbances by their distribution conditioned on the observations $\vec{o}$:

$$\vec{\mu}^o_\epsilon \triangleq \vec{\mu}_{\epsilon|o} = \vec{\mu}_\epsilon + \Sigma_{\epsilon,o}\Sigma_{o,o}^{-1}(\vec{o} - \vec{\mu}_o)$$

$$= \vec{\mu}_\epsilon + \Sigma_{\epsilon,\epsilon}S_o^t(S_o\Sigma_{\epsilon,\epsilon}S_o^t)^{-1}(\vec{o} - \vec{\mu}_o)$$

$$\Sigma^o_{\epsilon,\epsilon} \triangleq \Sigma_{\epsilon,\epsilon|o} = \Sigma_{\epsilon,\epsilon} - \Sigma_{\epsilon,o}\Sigma_{o,o}^{-1}\Sigma_{o,\epsilon}$$

$$= \Sigma_{\epsilon,\epsilon} - \Sigma_{\epsilon,\epsilon}S_o^t(S_o\Sigma_{\epsilon,\epsilon}S_o^t)^{-1}S_o\Sigma_{\epsilon,\epsilon}$$

where $S_o$ is the submatrix of $S$ containing all columns of $S$ but only those rows corresponding to the observed variables in $\vec{o}$.

We then evaluate the means $\vec{\mu}_{x^*|\hat{a}^*o}$ and variances $\Sigma_{x^*,x^*|\hat{a}^*o}$ of the variables in the counterfactual world ($\vec{X}^*$) under the action $\vec{a}$ using Eqs. (19) and (20), by replacing the prior distribution on the disturbances $\Sigma_{\epsilon_y,\epsilon_y}$ and $\mu_{\epsilon_y}$ with the posterior distribution $\Sigma^o_{\epsilon_y,\epsilon_y}$ and $\mu^o_{\epsilon_y}$:

$$\mu_{x^*|\hat{a}^*o} = \begin{bmatrix} (I - B_{yy})^{-1}(\vec{\mu}^o_{\epsilon_y} + B_{yz}a_z) \\ \vec{a}_z \end{bmatrix} (21)$$

$$\Sigma_{x^*,x^*|\hat{a}} = \begin{bmatrix} (I - B_{yy})^{-1}\Sigma^o_{\epsilon_y,\epsilon_y}((I - B_{yy})^{-1})^t & 0 \\ 0 & 0 \end{bmatrix} (22)$$

It is clear that this procedure can be applied to non-triangular matrices, as long as $S$ is nonsingular.

## 4    EXAMPLE

Consider the econometric structural equation model described in [Goldberger, 1992]:

$$q = b_1 p + d_1 i + u_1 \quad (23)$$
$$p = b_2 q + d_2 w + u_2 \quad (24)$$

where $q$ is the quantity of household demand for product A, $p$ is the unit price of product A, $i$ is household income, $w$ is wage rate for producing product A, $u_1$ is demand shock, and $u_2$ is supply shock.

We extend this model by incorporating an additional variable $r$, the household demand for some substitute product B, along with its structural equation

$$r = b_3 p + u_3$$

Let B stand for tea and A for coffee. Consider the following set of counterfactual queries:

1. Find the expected demand for coffee ($q$) had coffee prices ($p$) been controlled, say at $p = \$7.00$?

2. Find the expected demand for coffee ($q$) had coffee prices ($p$) been controlled, say at $p = \$7.00$, assuming the demand for tea subsequently reaches $r = 4$?

3. Given that the current demand for tea ($r$) is $r = 4$, find the expected demand for coffee ($q$) had coffee prices ($p$) been controlled, say at $p = 7.00$?

Note the difference between queries 2 and 3. Query 2 states that the price intervention occurs prior to our observation of product B's demand, while query 3 states that we first make an observation of product B's demand and then intervene to force product A's price.

The above counterfactual queries only involve the variables $\vec{X} = [P, Q, R]$; therefore, we may marginalize out all remaining variables in Eqs. (23) and (24), only retaining the distributions on $P$, $Q$, and $R$'s disturbance terms. Because $I$ and $W$ are exogenous (root) variables in the structural equations, we may combine $I$ and $U_1$ into one disturbance variable $\epsilon_q$. Likewise, $W$ and $U_2$ may be combined into one disturbance variable $\epsilon_p$. The structural equations for analyzing the above counterfactual queries may be reduced to

$$\vec{x} = B\vec{x} + \vec{\epsilon}$$
$$\begin{bmatrix} p \\ q \\ r \end{bmatrix} = \begin{bmatrix} 0 & b_2 & 0 \\ b_1 & 0 & 0 \\ b_3 & 0 & 0 \end{bmatrix} \begin{bmatrix} p \\ q \\ r \end{bmatrix} + \begin{bmatrix} \epsilon_p \\ \epsilon_q \\ \epsilon_r \end{bmatrix} \quad (25)$$

The causal structure for this model is shown in Figure 3.

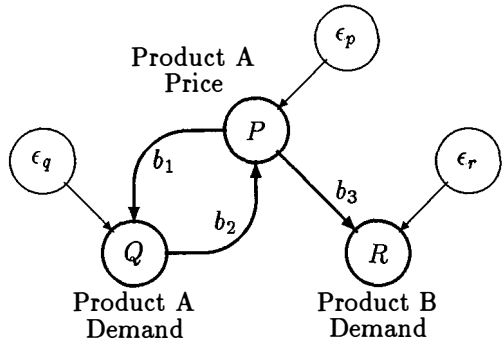

Figure 3: *Causal structure of an econometric model relating the demand for two products A and B and the price of product A. The variables are related according to the linear structural equations given in Eq. (25), where the disturbances $\epsilon_p$, $\epsilon_q$, and $\epsilon_r$ are independent and normally distributed.*

Because $R$ and $Q$ are d-separated ([Pearl, 1988]) by $P$ when the arrow $Q \longrightarrow P$ is removed, the observation of $R$ after $P$'s intervention has no impact on the evaluation of $Q$'s distribution. Therefore, the counterfactual distribution of demand for coffee $(Q)$ will be the same for queries 1 and 2.

Suppose that the parameters for this model are given by

$$B = \begin{bmatrix} 0 & 0.50 & 0 \\ -1.80 & 0 & 0 \\ 1.00 & 0 & 0 \end{bmatrix}$$
$$\vec{\mu}_\epsilon^t = [\ 0 \quad 19.00 \quad 3.00\ ]$$
$$\Sigma_{\epsilon,\epsilon} = \begin{bmatrix} 1.00 & 0 & 0 \\ 0 & 3.00 & 0 \\ 0 & 0 & 2.00 \end{bmatrix}$$

which reflects the following prior distribution on $\vec{X} = [P, Q, R]$:

$$\vec{\mu}_x^t = [\ 5.00 \quad 10.00 \quad 8.00\ ]$$
$$\Sigma_{x,x} = \begin{bmatrix} 0.48 & -0.08 & 0.48 \\ -0.08 & 1.73 & -0.08 \\ 0.48 & -0.08 & 2.48 \end{bmatrix}$$

The expected price of coffee is $5.00, while the average demand for coffee and tea are 10 units and 8 units, respectively.

Query 1 is interested in determining the distribution of demand for coffee $(Q)$, given that no observations have been made on the system, if we had intervened to force the price of coffee to $7.00. Evaluating the expressions in Eqs. (21) and (22), we obtain:

$$\vec{\mu}_{x^*|\hat{p}=7}^t = [\ 7.00 \quad 6.40 \quad 10.00\ ] \quad (26)$$
$$\sigma_{x^*|\hat{p}=7} = \begin{bmatrix} 0 & 0 & 0 \\ 0 & 3.00 & 0 \\ 0 & 0 & 2.00 \end{bmatrix}$$

We conclude that the average household demand for coffee and tea would be 6.4 units and 10 units, respectively, if the price of coffee were $7.00.

Query 3 asks for the expected demand demand for coffee $(Q)$ had the price of coffee been controlled at $7.00, given that demand for tea is currently 4 units. Applying the expressions in Eqs. (21) and (22):

$$\vec{\mu}_{x^*|\hat{p}=7,r=4}^t = [\ 7.00 \quad 5.13 \quad 6.78\ ] \quad (27)$$
$$\sigma_{x^*|\hat{p}=7,r=4} = \begin{bmatrix} 0 & 0 & 0 \\ 0 & 2.75 & -0.64 \\ 0 & -0.64 & 0.39 \end{bmatrix}$$

Note the importance of the observation of demand for tea $(R)$. In query 1, we found that forcing the price of coffee $(P)$ to $7.00 would reduce the expected demand for coffee $(Q)$ from 10 units to 6.4 units. The observation of a 4 unit demand for tea changes the expected demand for coffee to $\mu_{q|r=4} = 10.13$ units; if we intervene to force the price of coffee to $7.00, the expected demand for coffee $(Q)$ will be reduced from 10.13 to 5.13 units. Therefore, we see that enforcing a $7.00 price control on coffee would have a more adverse affect on the demand for coffee under the knowledge that the demand for tea was only 4 units. In addition, the expected demand for tea would increase to 6.78 units from the observed 4 units.

If we believe that the disturbance on the demand for coffee $(\epsilon_q)$ changes slowly, or at least changes infrequently, then we can use the results of this counterfactual distribution to determine whether price controls should now be imposed to meet our needs. In other words, the counterfactual distribution will tell us how we expect variables' distributions to change as a result of an external intervention applied in the present.

It is important to note the difference between counterfactual distributions (conditioned on observations and external intervention) and distributions simply conditioned on observations. Consider the distribution that would be computed from observing the price of coffee at $7.00 ($p = 7$) or from observing the demand for tea at 4 units and the coffee price at $7.00 ($r = 4, p = 7$):

$$\vec{\mu}_{x|p=7}^t = [\ 7.00 \quad 9.66 \quad 10.00\ ] \quad (28)$$
$$\sigma_{x,x|p=7} = \begin{bmatrix} 0 & 0 & 0 \\ 0 & 1.71 & 0 \\ 0 & 0 & 2.00 \end{bmatrix} \quad (29)$$
$$\vec{\mu}_{x|r=4,p=7}^t = [\ 7.00 \quad 9.66 \quad 4.00\ ] \quad (30)$$
$$\sigma_{x,x|r=4,p=7} = \begin{bmatrix} 0 & 0 & 0 \\ 0 & 1.71 & 0 \\ 0 & 0 & 0 \end{bmatrix} \quad (31)$$

Contrast the expected demand for coffee evaluated from these conditional distributions with that expected had the price of coffee been fixed by external intervention. In particular, compare Eq. (28) to Eq. (26) and Eq. (30) to Eq. (27). One reason it is incorrect to use distributions conditioned on observations for evaluating (economic) policies, is that such distributions convey false information about the post-intervention state of the disturbances. Accounting for the pre-intervention value of the controlled variables, which convey correct information about those disturbances, is important therefore for properly evaluating the effect of the intervention.

## 5 CONCLUSION

This paper has addressed the inadequacy of current techniques in econometrics and the social sciences for evaluating the potential effects of economic and social policies. Current techniques fail to correctly evaluate policies that control endogenous variables, that is, variables that are influenced by other variables in the system prior to enacting the policy.

We have addressed this deficiency by developing and applying a formalism for evaluating counterfactual conditionals in structural equation models. This method is applicable to the analysis of policies, even when the policy dictates intervention on an endogenous variable. An example was presented that demonstrates the disparity between analyses based on counterfactuals and reduced-form analysis which treats intervention as an observation on controlled variables.

The technique developed in this paper should also be applicable to AI problems in situations where a strategy is to be evaluated on the basis of structural equations with a given functional form. Examples are presented for causal models using boolean functions, with and without exceptions.

**Acknowledgements**

The research was partially supported by Air Force grant #AFOSR/F496209410173, NSF grant #IRI-9420306, and Rockwell/Northrop Micro grant #94-100.